\documentclass{article}

 \usepackage[dblblindworkshop, final]{neurips_2025}

\usepackage[utf8]{inputenc} 
\usepackage[T1]{fontenc}    
\usepackage{hyperref}       
\usepackage{url}            
\usepackage{booktabs}       
\usepackage{amsfonts}       
\usepackage{nicefrac}       
\usepackage{microtype}      
\usepackage[dvipsnames]{xcolor}         

\title{Video Finetuning Improves Reasoning Between Frames}
\workshoptitle{CogInterp: Interpreting Cognition in Deep Learning Models}
\usepackage{graphicx}
\usepackage{amsmath}
\usepackage{lineno}
\usepackage{subcaption}

\definecolor{darkblue}{rgb}{0, 0, 0.5}
\hypersetup{colorlinks=true, citecolor=darkblue, linkcolor=darkblue, urlcolor=darkblue}

\usepackage{comment}

\author{%
  Ruiqi Yang \\
  Brown University\\
  \texttt{ruiqi\_yang1@brown.edu} \\
  \And
  Tian Yun \\
  Brown University \\
  \texttt{tian\_yun@brown.edu} \\
  \And
  Zihan Wang \\
  Brown University \\
  \texttt{zihan\_wang3@brown.edu} \\
  \And
  Ellie Pavlick \\
  Brown University \\
  \texttt{ellie\_pavlick@brown.edu} \\
}

\begin{document}

\maketitle

\begin{abstract}

Multimodal large language models (LLMs) have made rapid progress in visual understanding, yet their extension from images to videos often reduces to a naive concatenation of frame tokens. 
In this work, we investigate what video finetuning brings to multimodal LLMs. We propose Visual Chain-of-Thought (vCoT), an explicit reasoning process that generates transitional event descriptions between consecutive frames. Using vCoT, we systematically compare image-only LVLMs with their video-finetuned counterparts, both with and without access to these transitional cues. Our experiments show that vCoT significantly improves the performance of image-only models on long-form video question answering, while yielding only marginal gains for video-finetuned models—suggesting that the latter already capture frame-to-frame transitions implicitly. Moreover, we find that video models transfer this temporal reasoning ability to purely static settings, outperforming image models' baselines on relational visual reasoning tasks.

\end{abstract}

\section{Introduction}

Multimodal large language models (LLMs) have reached remarkable progress in understanding visual contents through the integration of pretrained LLMs with pretrained image or video encoders. 
Image LLMs, such as \citep{li2023blip} and \citep{liu2023visual}, have demonstrated strong capabilities on various downstream tasks, such as image captioning \citep{chen2015microsoft, plummer2015flickr30k}, tabular data understanding \citep{chen2019tabfact}, visual question answering \citep{bigham2010vizwiz,goyal2017making,hudson2019gqa}. However, their naive extension to the video domain often involves frame-by-frame tokenization without true temporal understanding. Consequently, these models tend to rely on superficial visual cues and struggle when a task requires reasoning over implicit transitions in between multiple video frames.

In contrast, video LLMs are designed to reach better video understanding by additional finetuning on video data and more inductive biases, such as additional temporal positional encoding with RoPE \citep{bai2025qwen2, hong2025glm}. Despite their architectural advantages, pretrained video LLMs often underperform on tasks requiring deep temporal reasoning unless explicitly fine-tuned on temporally grounded datasets \citep{gao2017tall, liu2024tempcompass, ren2024timechat}. This raises a critical question: To what extent does video-based fine-tuning enhance the reasoning capabilities of LLMs beyond what image-based models can achieve?

In this work, we investigate whether the video finetuning of video LLMs improves their understanding and reasoning between frames. To approach this question, we systematically comparing video LLMs with their image LLM counterparts, which share the same architecture and image training data, and differ with the video LLMs mainly in terms of the absense of video finetuning. We introduce a visual Chain-of-Though (vCoT) to obtain explicit descriptions in between each pair of frames to measure models' performance with and without these descriptions on EgoSchema \citep{mangalam2023egoschema}, which is a long-form video understanding benchmark and cannot be simply solved by observing static images. Specifically, we make the following contributions:
\begin{enumerate}
    \item We consider pairs of image LLMs and video LLMs and measure models' performance with and without vCoT. We demonstrate that video LLMs gain marginal benefits with explicit in-between frame descriptions, while image LLMs gain substantial improvement, reflecting that video LLMs may learn to infer the frame transition implicitly. 
    \item We randomly sample videos from other samples and replace the video stream or text infill stream to examine the robustness to noise of video LLMs and image LLMs. We observe that video LLMs are more robust noise from either modality.
    \item We study whether the reasoning capabilities between frames can be extended from video domain to static images. We focus on RAVEN \citep{zhang2019raven}, and observe video LLMs outperform image LLMs on all sub-tasks in RAVEN, showing that video LLMs achieve better reasoning between not only the contiguous image frames but also the static images.
\end{enumerate}

\begin{figure}[t]
  \centering
  \includegraphics[width=1\linewidth]{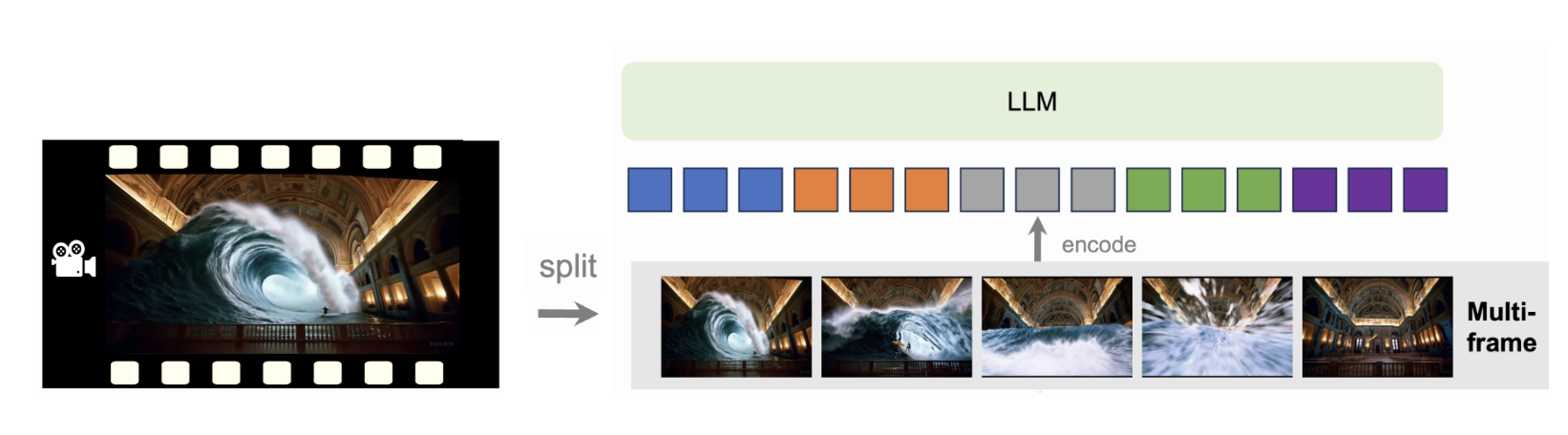}
  \caption{Video LLMs typically sample a number of frames from a video and digest them as a sequence of concatenated visual tokens \citep{zhang2024llavanextvideo}.}
  \label{fig:videollm}
\end{figure}

\begin{figure}[t]
  \centering
  \includegraphics[width=\linewidth]{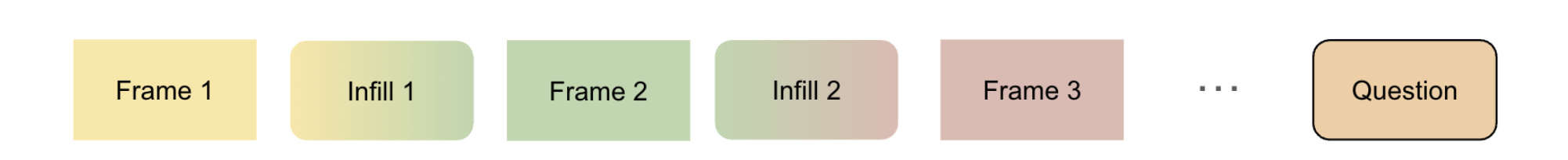}
  \caption{Visual CoT prompting pipeline. For each clip, transitional text infills are generated between every adjacent frame pair, interleaved with corresponding frames, and followed by the downstream question.}
  \label{fig:cot_pipeline}
\end{figure}

\section{Method}
\label{sec:method}

To study whether video LLMs achieve stronger inter-frame reasoning than image-based counterparts, we introduce \textbf{Visual Chain-of-Thought (vCoT)} infills—explicit textual reasoning inserted between consecutive frames. We investigate whether these infills improve the performance of (i) an image-only vision–language model (VLM) and (ii) its video-finetuned counterpart. If adding vCoT substantially boosts accuracy, it suggests that the model benefits from explicit reasoning and lacks implicit temporal understanding. Conversely, minimal or negative gains imply that the model already reasons over frames without additional cues.

\subsection{Visual Chain-of-Thought (vCoT) Infills}
\label{sec:method:cot}

\begin{figure}[t]
  \centering
  \includegraphics[width=\linewidth]{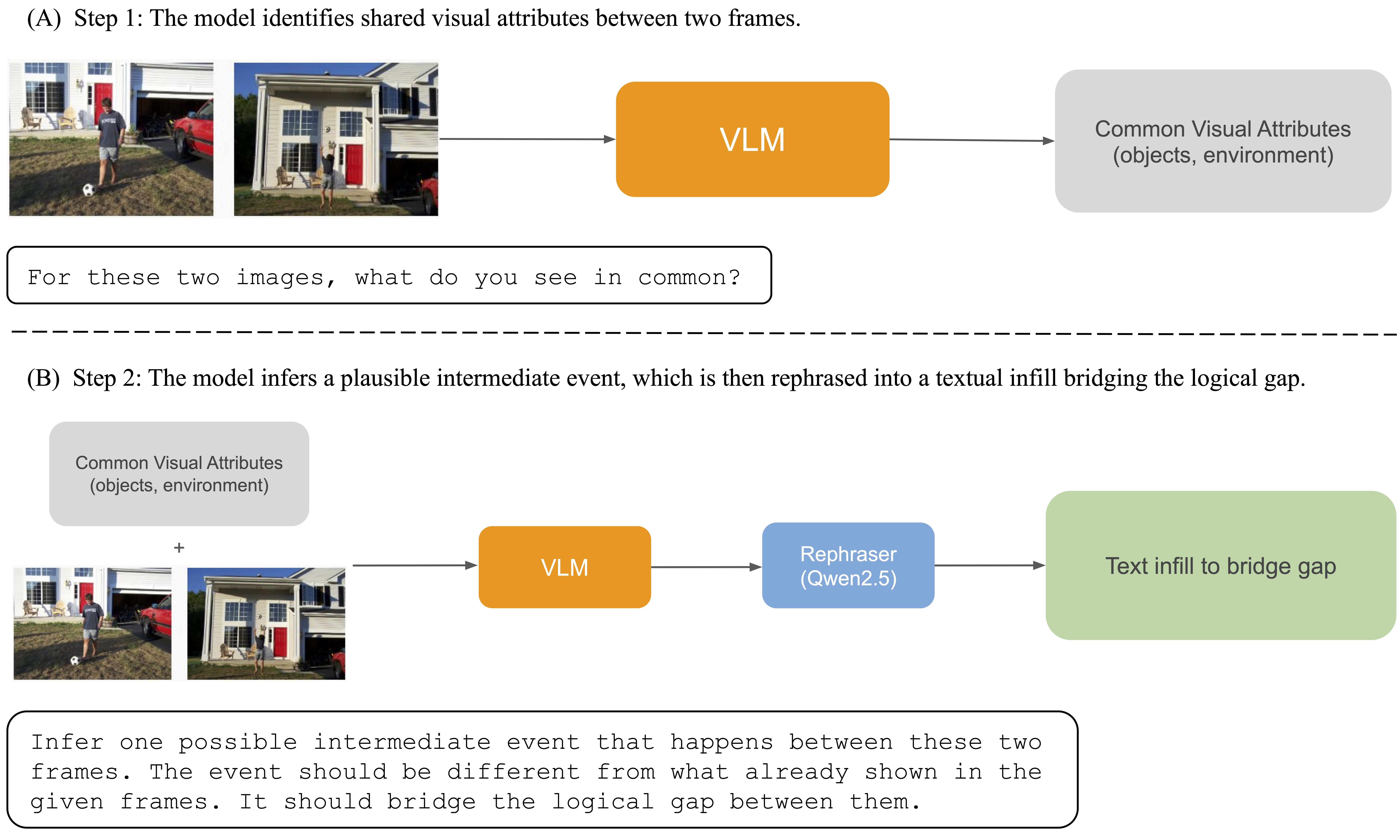}
  \caption{Visual CoT for inter-frame reasoning. 
  Step 1: The model identifies shared visual attributes between two frames. 
  Step 2: It infers a plausible intermediate event, which is then rephrased into a textual infill bridging the logical gap.}
  \label{fig:interframe_reasoning}
\end{figure}

Motivated by chain-of-thought prompting in NLP \citep{wei2022chain}, we introduce \textbf{visual Chain-of-Thought} (\textsc{vCoT})—a two-step reasoning process that first captures the static context shared by two frames and then infers a plausible event connecting them. Each one-sentence \emph{transitional caption} explicitly describes how the scene evolves between consecutive frames $(F_i, F_{i+1})$, serving as a \emph{text infill} interleaved with the original video sequence. These infills provide interpretable intermediate steps that make temporal reasoning explicit and enhance the model’s understanding of frame-to-frame continuity. The vCoT generation process consists of two sequential prompts (Figure~\ref{fig:interframe_reasoning}):

\paragraph{Step 1: Common Visual Attributes.}
We first query the model: \textit{“For these two images, what do you see in common?”}  
This encourages the model to identify shared scene elements—such as objects, background, or spatial configuration—providing a stable context across frames.

\paragraph{Step 2: Bridging Event Inference.}
Next, using both frames and the identified context, we prompt:  
\textit{“Infer one possible intermediate event that happens between these two frames. The event should be different from what is already shown and should bridge the logical gap between them.”}  
This step yields a plausible, one-sentence description of the temporal transition (e.g., “the person kicks the ball toward the house”). To maintain clarity and conciseness, each event description is then rephrased using a lightweight Qwen-2.5 model \citep{yang2024qwen2}.

For each video clip, the above process is applied iteratively to every consecutive frame pair, producing a chain of inferred events. The resulting interleaved sequence—frames and infills—is then combined with the task question (multiple-choice or open-ended) and passed to the VLM for final prediction.

\begin{table}[t]
  \centering
  \footnotesize
  \caption{Controlled experiment results on the EgoSchema benchmark.  
  (a) LLaVA models and (b) InternVL models evaluated with and without vCoT.}
  \begin{subtable}[t]{0.48\linewidth}
    \centering
    \caption{LLaVA variants. Accuracy (\%).}
    \begin{tabular}{lccc}
      \toprule
      \textbf{Model} & \textbf{\#F} & \textbf{Base} & \textbf{+vCoT} \\
      \midrule
      LLaVA-NeXT & 5 & 44.0 & 51.4 \color{ForestGreen}{\scriptsize(+7.4)} \\
      LLaVA-NeXT-Video & 5 & 47.0 & 48.6 \color{ForestGreen}{\scriptsize(+1.6)} \\
      LLaVA-NeXT & 10 & 49.2 & 55.4 \color{ForestGreen}{\scriptsize(+6.2)} \\
      LLaVA-NeXT-Video & 10 & 49.0 & 51.4 \color{ForestGreen}{\scriptsize(+2.4)} \\
      \bottomrule
    \end{tabular}
    \label{tab:egoschema_a}
  \end{subtable}
  \hfill
  \begin{subtable}[t]{0.48\linewidth}
    \centering
    \caption{InternVL variants. Accuracy (\%).}
    \begin{tabular}{lccc}
      \toprule
      \textbf{Model} & \textbf{\#F} & \textbf{Base} & \textbf{+vCoT} \\
      \midrule
      InternVL-Image & 5 & 38.4 & 40.4 \color{ForestGreen}{\scriptsize(+2.0)} \\
      InternVL-Video & 5 & 44.6 & 42.4 \color{red}{\scriptsize(–2.2)} \\
      InternVL-Image & 10 & 37.4 & 42.6 \color{ForestGreen}{\scriptsize(+5.2)} \\
      InternVL-Video & 10 & 45.8 & 49.0 \color{ForestGreen}{\scriptsize(+3.2)} \\
      \bottomrule
    \end{tabular}
    \label{tab:egoschema_b}
  \end{subtable}
  \label{tab:egoschema_combined}
\end{table}

\section{Results}
\label{sec:results}

We evaluate vCoT on two representative LVLMs, \textsc{LLaVA-NeXT} \citep{liu2024llavanext} and \textsc{InternVL} \citep{chen2024far}, along with their video-finetuned counterparts, \textsc{LLaVA-NeXT-Video} \citep{zhang2024llavanextvideo} and \textsc{InternVL2} \citep{chen2024expanding}. Within each pair, the vision encoder, language backbone, and cross-modal projector remain identical; the only difference lies in whether video finetuning is applied. This controlled setup ensures that any observed performance variation can be attributed specifically to the effect of video finetuning.

Building on this evaluation design, we report three sets of findings.
First, we conduct a controlled study on the EgoSchema long-form video-QA benchmark \citep{mangalam2023egoschema}, highlighting the impact of vCoT under matched model capacity and data conditions.
Second, to examine modality sensitivity, we introduce a shuffling diagnostic that deliberately introduces conflicting visual or textual evidence.
Finally, we assess the transferability of vCoT to a different reasoning domain, the \textsc{i-Raven} logical-reasoning suite.

\subsection{Controlled experiment on EgoSchema}
\label{sec:egoschema}

EgoSchema subset comprises $500$ three‑minute egocentric videos paired with one
multiple‑choice question each.  
Following VLMEvalKit~\citep{duan2024vlmevalkit}, we prompt the model to
output an option index (``A–D'') and use
\textsc{Qwen2.5‑7B‑Chat} \citep{yang2024qwen2} as a judge.
All prompts replicate the official VLMEvalKit template.

Table~\ref{tab:egoschema_combined} shows LLaVA and InternVL models' performance on EgoSchema with and without vCoT. Our experiments reveal that explicit temporal reasoning through vCoT consistently enhances image models' performance across all tested configurations, with improvements ranging from $+1.6\%$ to $+7.4\%$ over the frame-only baseline. This underscores the importance of structured temporal understanding for long-form video comprehension.

Notably, the benefits of vCoT are most pronounced in models that lack prior video supervision. For instance, when using dense temporal sampling (\#F=5), the image-only \textsc{LLaVA‑NeXT} model sees a substantial performance gain of $+7.4\%$, compared to only $+1.6\%$ for its video-finetuned counterpart. A similar trend is observed across the InternVL variants. This discrepancy suggests that vCoT infills serve as a crucial supplementary reasoning mechanism for models not pretrained on video data, while models already exposed to video supervision may implicitly capture transitional dynamics, thereby reducing—or even negating—the marginal utility of vCoT.

\subsection{Modality‑shuffling ablations}
\label{sec:shuffling}
To disentangle reliance on visual versus textual cues, we construct two perturbations (Figure~\ref{fig:shuffle}): (1) \textit{Visual shuffle}: replace every video frame with a frame from an unrelated clip while keeping the text infills intact. (2) \textit{Text shuffle}: keep the original frames but swap the text infills with those from a different video.

\begin{table}[t]
  \centering
  \caption{vCoT accuracy and degradation on the EgoSchema benchmark under two perturbation strategies: visual shuffle and text shuffle.}
  \footnotesize
  \setlength{\tabcolsep}{5pt}
  \label{tab:replacement_vcot_combined}

  \begin{tabular}{lccccc}
    \toprule
    \textbf{Model ID} & \textbf{\#F} & \textbf{vCoT} 
    & \textbf{visual shuffle} & \textbf{text shuffle} \\
    \midrule
    LLaVA‑NeXT         & 5  & 51.4 & 39.8 \color{red}{\scriptsize(–11.6)} & 42.0 \color{red}{\scriptsize(–9.4)} \\
    LLaVA‑NeXT‑Video   & 5  & 48.6 & 41.6 \color{red}{\scriptsize(–7.0)}  & 47.0 \color{red}{\scriptsize(–1.6)} \\
    LLaVA‑NeXT         & 10 & 55.4 & 51.8 \color{red}{\scriptsize(–3.6)}  & 45.0 \color{red}{\scriptsize(–10.4)} \\
    LLaVA‑NeXT‑Video   & 10 & 51.4 & 46.4 \color{red}{\scriptsize(–5.0)}  & 45.4 \color{red}{\scriptsize(–6.0)} \\
    \bottomrule
  \end{tabular}

\end{table}

\begin{table}[t]
  \caption{Accuracy (\%) on the i-RAVEN benchmark, comparing model performance across different reasoning rule types. The left block summarizes accuracy for position-based rules (object alignment and spatial distribution), while the right block evaluates relational and directional rules (object interactions and spatial reasoning).
  }
  \centering
  \footnotesize
  \setlength{\tabcolsep}{4pt}
  \label{tab:raven_combined}
  
  \begin{tabular}{l|ccc|cccc|c}
    \toprule
    \textbf{Model ID} 
    & \textbf{center} & \textbf{dist\_4} & \textbf{dist\_9} & \textbf{in/out} & {\small\textbf{indist4/out}} & \textbf{L/R} & \textbf{U/D} & \textbf{Avg.} \\
    \midrule
    InternVL‑Image  
    & 14.8 & 14.4 & 15.2 
    & 11.6 & 13.2 & \textbf{15.2} & \textbf{14.4} & 14.1 \\
    
    InternVL‑Video  
    & \textbf{15.6} & \textbf{16.0} & \textbf{15.8} 
    & \textbf{13.8} & \textbf{17.0} & 14.0 & 14.2 & \textbf{15.2} \\
    
    \midrule
    LLaVA‑Image     
    & 7.0 & 8.0 & 15.0 
    & 7.0 & 9.0 & 12.0 & 14.0 & 10.3 \\
    
    LLaVA‑Video     
    & 7.0 & \textbf{14.0} & \textbf{16.0} 
    & \textbf{8.0} & \textbf{13.0} & \textbf{14.0} & \textbf{21.0} & \textbf{13.3} \\
    \bottomrule
  \end{tabular}
\end{table}

Table~\ref{tab:replacement_vcot_combined} shows the results. Both image-based and video-based LLaVA models are sensitive to visual perturbations; however, the video variant exhibits significantly greater robustness to textual noise, suggesting a stronger reliance on the visual modality.

\subsection{Transferability to Relational Visual Reasoning}
\label{sec:raven}
A natural question is whether the ability to reason across video frames also transfers to reasoning across static image frames. To examine this, we turn to the \textsc{i-Raven} benchmark \cite{hu2021stratified}, a relational visual reasoning suite derived from Progressive Matrices. In this task, models must infer abstract visual rules from a set of panels and apply them to identify the correct completion, making it a challenging test of non-temporal relational inference.

Despite the absence of explicit temporal signals in the task, models finetuned on video data consistently outperform their image-only counterparts, indicating that temporal reasoning can transfer to relational reasoning. Specifically, video-finetuned InternVL and LLaVA models yield overall gains of $+1.1\%$ and $+3.0\%$, respectively, compared to their non-video baselines. The most notable improvements occur in categories involving spatial layouts and relative positioning. For example, InternVL shows a $+3.8\%$ gain on the \texttt{indist4/out} rule, while LLaVA achieves a $+7.0\%$ improvement on \texttt{U/D}. These results suggest that video finetuning may induce a stronger inductive bias toward relational structures, enabling models to better capture abstract spatial dependencies even in tasks devoid of temporal context.

\section{Conclusion}
In this paper, we investigate the unique benefits that video finetuning brings to multimodal LLMs through a controlled comparison between image-based models and their video-finetuned counterparts. To this end, we introduce Visual Chain-of-Thought (vCoT), which generates explicit textual infills between consecutive frames to represent intermediate reasoning steps.  Through controlled experiments between image LLMs and their video counterparts, we demonstrate that video-finetuned models already perform implicit inter-frame reasoning, and that this capability naturally transfers to static visual relational reasoning tasks.

\section{Acknowledgement}
We would like to thank all reviewers and the area chair for their valuable feedback. 
We would like to thank Zitian Tang, Shijie Wang, and other members of the SuperLab at Brown University for their discussions and insights. The project depicted is sponsored in part by a Young Faculty Award from the Defense Advanced Research Projects Agency, Grant \#D24AP00261. The content of the information does not necessarily reflect the position, or the policy of the government and no official endorsement of this work should be inferred.

\bibliographystyle{plainnat}
\bibliography{custom}

\newpage
\appendix

\section{Appendix}
\label{sec:appendix}
\subsection{Modality shuffling demonstration}
\begin{figure}[h]
  \centering
  \includegraphics[width=1\linewidth]{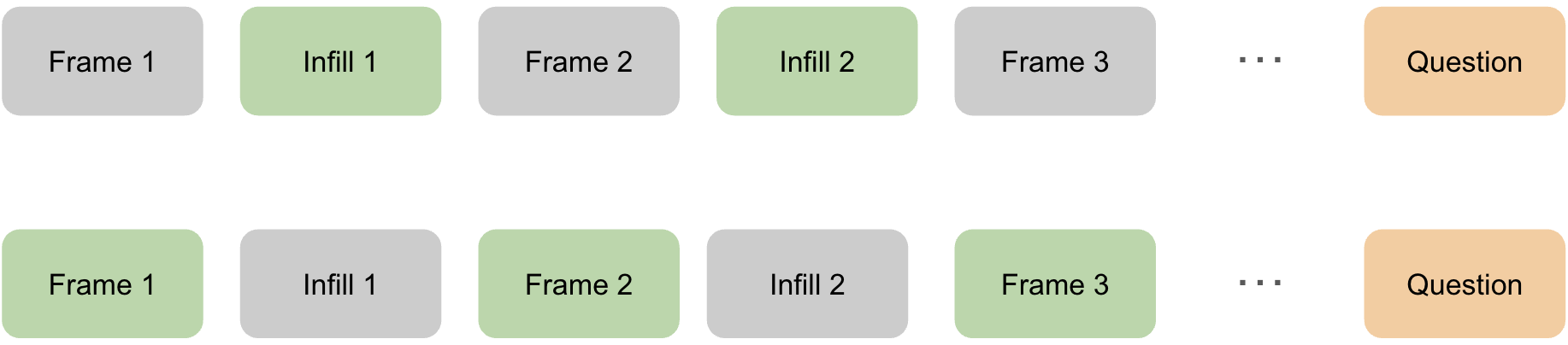}
  \caption{Shuffling to create conflicting modalities. Grey = retained modality; green = shuffled modality. Top row = text shuffle; bottom row = visual shuffle.}
  \label{fig:shuffle}
\end{figure}

\subsection{Frame captions as text infills}
To distinguish vCoT infills from simple frame captions, we evaluate LLaVA models on EgoSchema using image captions as inter-frame text. Results show that while captions provide some benefit, vCoT infills consistently yield greater improvements, offering stronger guidance for temporal reasoning (see Table \ref{tab:egoschema_c}).

\begin{table}[h]
\centering
\caption{Accuracy (\%) with frame captions as vCoT on LLaVA variants.}
\begin{tabular}{lcccc}
  \toprule
  \textbf{Model ID} & \textbf{\#F} & \textbf{Baseline} & \textbf{Captions} & \textbf{vCoT} \\
  \midrule
  LLaVA-NeXT & 5  & 44.0 & 48.2 & 51.4 \color{ForestGreen}{\scriptsize(+3.2)} \\
  LLaVA-NeXT‑Video & 5 & 47.0 & 45.0 & 48.6 \color{ForestGreen}{\scriptsize(+3.6)} \\
  LLaVA-NeXT & 10 & 49.2 & 54.6 & 55.4 \color{ForestGreen}{\scriptsize(+0.8)} \\
  LLaVA-NeXT‑Video & 10 & 49.0 & 50.0 & 51.4 \color{ForestGreen}{\scriptsize(+1.4)} \\
  \bottomrule
\end{tabular}
\label{tab:egoschema_c}
\end{table}

\subsection{LoRA fine‑tuning to neutralise data‑scale effects.}
The video-finetuned variant of \textsc{LLaVA‑NeXT} is trained on approximately \(100\text{k}\) additional video‑instruction pairs beyond the \(600\text{k}\) image prompts used for the base model \citep{zhang2024llavanextvideo}. To disentangle the effect of data scaling, we perform parameter‑efficient LoRA fine‑tuning on image LLaVA-NeXT. Specifically, we train rank‑128 LoRA \citep{hu2022lora} adapters on \(100\text{k}\) high‑quality image instructions from \texttt{ShareGPT‑4V} \citep{cui2025comprehensive} and \texttt{ShareGPT‑4o}, so that the total amount of training data matches that of the video variant.

\begin{table}[ht]
  \centering
  \footnotesize
    \caption{Accuracy (\%) on the EgoSchema subset ($n{=}500$) \emph{after}
           LoRA fine‑tuning the image‑only backbone.}
  \begin{tabular}{lccccc}
    \toprule
    \textbf{Model ID} & \textbf{\#F} & \textbf{Stride}
      & \textbf{Baseline} & \textbf{vCoT}\\
    \midrule
    LLaVA‑NeXT (LoRA) & 5  & 1 & 45.8 & 45.0 \color{red}{\scriptsize(–0.8)} \\
    LLaVA‑NeXT‑Video  & 5  & 1 & 47.0 & 48.6 \color{ForestGreen}{\scriptsize(+1.6)} \\
    LLaVA‑NeXT (LoRA) & 5  & 2 & 36.2 & 37.4 \color{ForestGreen}{\scriptsize(+1.2)} \\
    LLaVA‑NeXT‑Video  & 5  & 2 & 36.8 & 42.2 \color{ForestGreen}{\scriptsize(+5.4)} \\
    \midrule
    LLaVA‑NeXT (LoRA) &10 & 1 & 49.2 & 55.3 \color{ForestGreen}{\scriptsize(+6.1)} \\
    LLaVA‑NeXT‑Video  &10 & 1 & 49.0 & 51.4 \color{ForestGreen}{\scriptsize(+2.4)} \\
    LLaVA‑NeXT (LoRA) &10 & 2 & 39.8 & 43.0 \color{ForestGreen}{\scriptsize(+3.2)} \\
    LLaVA‑NeXT‑Video  &10 & 2 & 39.8 & 45.4 \color{ForestGreen}{\scriptsize(+5.6)} \\
    \bottomrule
  \end{tabular}
  \label{tab:egoschema_lora}
\end{table}

After LoRA tuning, the earlier trend—where vCoT consistently enhanced performance, particularly for image-only models—begins to shift significantly (see Table \ref{tab:egoschema_lora}). In particular, with dense temporal sampling (\#F=5, stride=1), the vCoT module no longer provides gains for the image-only model—in fact, performance drops by $0.8\%$. This reversal suggests that the earlier advantage of vCoT in image-based models, relative to video-finetuned models, may have been driven by differences in training data scale, rather than by an inherent deficiency in temporal understanding. Thus, the emergence of implicit temporal reasoning might not be exclusive to video finetuning. 

Additionally, a second phenomenon emerges: potential degradation in textual modeling. Prior work has shown that EgoSchema questions often rely on shortcut textual cues, such as those derived from image captions~\citep{zhang2023simple, wang2024videoagent}. LoRA adaptation on 100k image-instruction pairs may induce forgetting, weakening the model’s capacity to reason with such cues. This degradation in textual understanding likely contributes to the reduced effectiveness of vCoT-generated text infills in post-LoRA settings.

\subsection{Limitations}
The main limitation of this project is the inability to train control models from scratch.
Ideally, we would use a small LLM (e.g., 0.5B parameters) and train two variants: one on image data and the other on video data.
This setup would allow us to fully control the training pipeline and ablate the effects of various inductive biases—such as the use of temporal encoders, temporal positional embeddings, and other architectural components.

\vspace{0.5em}
\noindent
Additionally, greater care must be taken in selecting which inductive biases to control in such experiments.
Interestingly, our results show that LLaVA-NeXT often outperforms LLaVA-NeXT-Video, which is unexpected and suggests that the latter may not serve as a strong representative of video LLMs due to its limited performance.
We argue that all inductive biases—aside from those inherent to the base LLM architecture—should be incorporated if they enhance the model’s ability to learn from image or video data.
This approach would allow us to obtain upper-bound variants of both image- and video-trained LLMs starting from the same initialization, enabling a fair and meaningful comparison.
\end{document}